\pdfoutput=1

\documentclass[11pt]{article}

\usepackage{ACL2023}

\usepackage{times}
\usepackage{latexsym}

\usepackage[T1]{fontenc}

\usepackage[utf8]{inputenc}

\usepackage{microtype}

\usepackage{inconsolata}

\usepackage{multirow}
\usepackage{multicol}
\usepackage{makecell}

\usepackage{times}
\usepackage{latexsym}
\usepackage{graphicx}
\usepackage{subfigure}
\usepackage{float}
\graphicspath{{images}}
\usepackage{stfloats}
\usepackage{booktabs}
\usepackage{indentfirst}
\usepackage{amsmath}
\usepackage{amssymb}

\usepackage{enumitem}

\usepackage{colortbl}
\usepackage{arydshln}

\usepackage[linesnumbered,ruled,vlined]{algorithm2e}

\SetCommentSty{mycommfont}

\SetKwInput{KwInput}{Input}                
\SetKwInput{KwOutput}{Output}              

\usepackage{CJKutf8}

\usepackage{listings}
\usepackage{xcolor}

\definecolor{codegreen}{rgb}{0,0.6,0}
\definecolor{codegray}{rgb}{0.5,0.5,0.5}
\definecolor{codepurple}{rgb}{0.58,0,0.82}
\definecolor{backcolour}{rgb}{0.95,0.95,0.92}

\lstdefinestyle{mystyle}{
    backgroundcolor=\color{backcolour},   
    commentstyle=\color{codegreen},
    keywordstyle=\color{magenta},
    numberstyle=\tiny\color{codegray},
    stringstyle=\color{codepurple},
    basicstyle=\ttfamily\small,
    breakatwhitespace=false,         
    breaklines=true,                 
    captionpos=b,                    
    keepspaces=true,                 
    numbers=left,                    
    numbersep=5pt,                  
    showspaces=false,                
    showstringspaces=false,
    showtabs=false,                  
    tabsize=2
}

\title{AMAS: Adaptively Determining Communication Topology for LLM-based Multi-Agent System}

\author{
    Hui Yi Leong\textsuperscript{1,}\thanks{\ \ These authors contribute equally.}, Yuheng Li\textsuperscript{2,$*$}, Yuqing Wu\textsuperscript{1,$*$}, Wenwen Ouyang\textsuperscript{3}, \\
    \textbf{Wei Zhu\textsuperscript{4,}\thanks{\ \ Corresponding author. For any inquiries, please contact: michaelwzhu91@gmail.com, jiechao@stanford.edu. }, Jiechao Gao\textsuperscript{5,$\dagger$}}, \textbf{Wei Han\textsuperscript{6}} \\
    \small \textsuperscript{1}University of Chicage, Chicago, IL, United States. \ \ \ \small \textsuperscript{2}Johns Hopkins University, Baltimore, MD, United States \\
    \small \textsuperscript{3}Carnegie Mellon University, Pittsburgh, PA, United States \ \ \ \small \textsuperscript{4}University of Hong Kong, Hong Kong, HK, China \\
    \small \textsuperscript{5}Stanford University, Stanford, CA, United States \\
    \small \textsuperscript{6} Independent Researcher, Austin, TX, United States. Email: palebluedot.milkyway@gmail.com
}

\begin{document}
\maketitle


\begin{abstract}

Although large language models (LLMs) have revolutionized natural language processing capabilities, their practical implementation as autonomous multi-agent systems (MAS) for industrial problem-solving encounters persistent barriers. Conventional MAS architectures are fundamentally restricted by inflexible, hand-crafted graph topologies that lack contextual responsiveness, resulting in diminished efficacy across varied academic and commercial workloads. To surmount these constraints, we introduce AMAS, a paradigm-shifting framework that redefines LLM-based MAS through a novel dynamic graph designer. This component autonomously identifies task-specific optimal graph configurations via lightweight LLM adaptation, eliminating the reliance on monolithic, universally applied structural templates. Instead, AMAS exploits the intrinsic properties of individual inputs to intelligently direct query trajectories through task-optimized agent pathways. Rigorous validation across question answering, mathematical deduction, and code generation benchmarks confirms that AMAS systematically exceeds state-of-the-art single-agent and multi-agent approaches across diverse LLM architectures. Our investigation establishes that context-sensitive structural adaptability constitutes a foundational requirement for high-performance LLM MAS deployments.

\end{abstract}

\begin{CJK*}{UTF8}{gbsn}

\section{Introduction}

Despite achieving unprecedented success in natural language processing benchmarks, large language models (LLMs) have established SOTA performance across specialized domains including domain-specific question answering, mathematical deduction, safety alignment, and instruction comprehension \cite{qin2023chatgpt,PromptCBLUE,text2dt_shared_task,Text2dt,zhu_etal_2021_paht,Li2023UnifiedDR,Zhu2023BADGESU,Zhang2023LECOIE,Zhu2023OverviewOT,guo-etal-2021-global,zhu-etal-2021-discovering,Zheng2023CandidateSF,info:doi/10.2196/17653,Zhang2023NAGNERAU,Zhang2023FastNERSU,Wang2023MultitaskEL,Zhu2019TheDS,Zhu2021LeeBERTLE,Zhang2021AutomaticSN,Wang2020MiningIH}. However, their transition from sophisticated language processors to autonomous problem-solving engines remains fraught with unresolved challenges \cite{huang2023c,li2023cmmlu,Cui2023UltraFeedbackBL,wang2024ts,yue2023-TCMEB,Zhang2023LearnedAA,2023arXiv230318223Z,Xu2023ParameterEfficientFM,Ding2022DeltaTA,Xin2024ParameterEfficientFF}. Current research primarily focuses on single-model applications, leaving critical gaps in operational deployment frameworks for LLM-driven agent ecosystems. Notably, the momentum toward LLM-based agent architectures has accelerated dramatically, with both industrial practitioners and academic researchers increasingly prioritizing this paradigm shift for scalable problem-solving architectures.

A vibrant scholarly pursuit has centered on architecting LLM-driven agent frameworks, evolving from foundational GPT-3 \cite{brown2020language} through sophisticated few-shot prompting mechanisms that harness LLMs' intrinsic in-context learning potential. Current single-agent implementations increasingly deploy structured reasoning protocols—such as Chain of Thought (COT) \cite{Wei2022ChainOT}, ReAct \cite{yao2022react}, Tree of Thought (TOT) \cite{muralidharan2024deliberate}, Reflexion \cite{shinn2024reflexion}, and Graph of Thought (GOT) \cite{besta2024graph}—to elevate textual reasoning efficacy. To transcend these boundaries, LLM-powered multi-agent systems (MAS) \cite{zeng2022socratic,zhuge2024gptswarm,li2023camel} have gained traction across industrial and academic domains. These systems deploy multiple LLM instances with distinct functional roles \cite{park2023generative}, enabling natural language coordination to collectively resolve complex problems. This distributed intelligence paradigm consistently surpasses single-agent benchmarks by leveraging specialized agent expertise and emergent collective cognition \cite{minsky1988society}. Nevertheless, prevailing MAS approaches persistently rely on handcrafted agent collaboration topologies. GPTSwarm \cite{zhuge2024gptswarm} represents a pivotal departure by formalizing MAS as a parameterized graph, employing reinforcement learning (RL) to autonomously refine structural configurations for optimal task execution.

This study introduces the Adaptive Multi-Agent System (AMAS) architecture to overcome fundamental constraints in contemporary methodologies \cite{zhu2024iapt,zhu-tan-2023-spt,Liu2022FewShotPF,xie2024pedro,Cui2023UltraFeedbackBL,zheng2024nat4at,zhu2023acf,gao2023f,zuo-etal-2022-continually,zhang-etal-2022-pcee,sun-etal-2022-simple,zhu-etal-2021-gaml,Zhu2021MVPBERTMP,li-etal-2019-pingan,zhu2019panlp,zhu2019dr,zhou2019analysis,zhang2025time,wang2025ts,liu2025parameter,yi2024drum,tian2024fanlora}. Preliminary empirical analysis uncovers pronounced heterogeneity across task-specific samples, demonstrating that no singular graph topology consistently achieves optimal outcomes within the MAS paradigm. Instead, numerous graph configurations yield performance metrics that closely approximate the highest-performing variant. This empirical insight propels a paradigm shift: rather than enforcing a static graph architecture, we develop a context-aware designer that autonomously adapts to input characteristics. The AMAS framework harnesses parameter-efficient adaptation protocols for large language models to construct this designer, enabling real-time graph selection tailored to each individual sample without task-specific reconfiguration.

This research establishes a rigorous empirical validation of the AMAS architecture across heterogeneous workloads, encompassing open-domain question answering, formal mathematical reasoning, and program synthesis challenges. Across all evaluated scenarios—regardless of underlying LLM architecture—the framework systematically outperforms both monolithic agents and conventional multi-agent baselines. Robust experimental evidence confirms the framework's cross-domain adaptability and operational robustness. Our key innovations are articulated as follows:  
\begin{itemize}  
\item We refine graph topology quality through adaptive integration of actor-critic dynamics within reinforcement learning-driven optimization pipelines.  
\item The architecture implements a dynamic graph selection mechanism that autonomously determines optimal structural configuration from candidate ensembles upon sample ingestion.  
\item Extensive empirical validation and mechanistic analysis substantiate AMAS's superior task resolution efficacy compared to state-of-the-art multi-agent paradigms.  
\end{itemize}

\begin{figure*}
\centering
\includegraphics[width=0.86\textwidth]{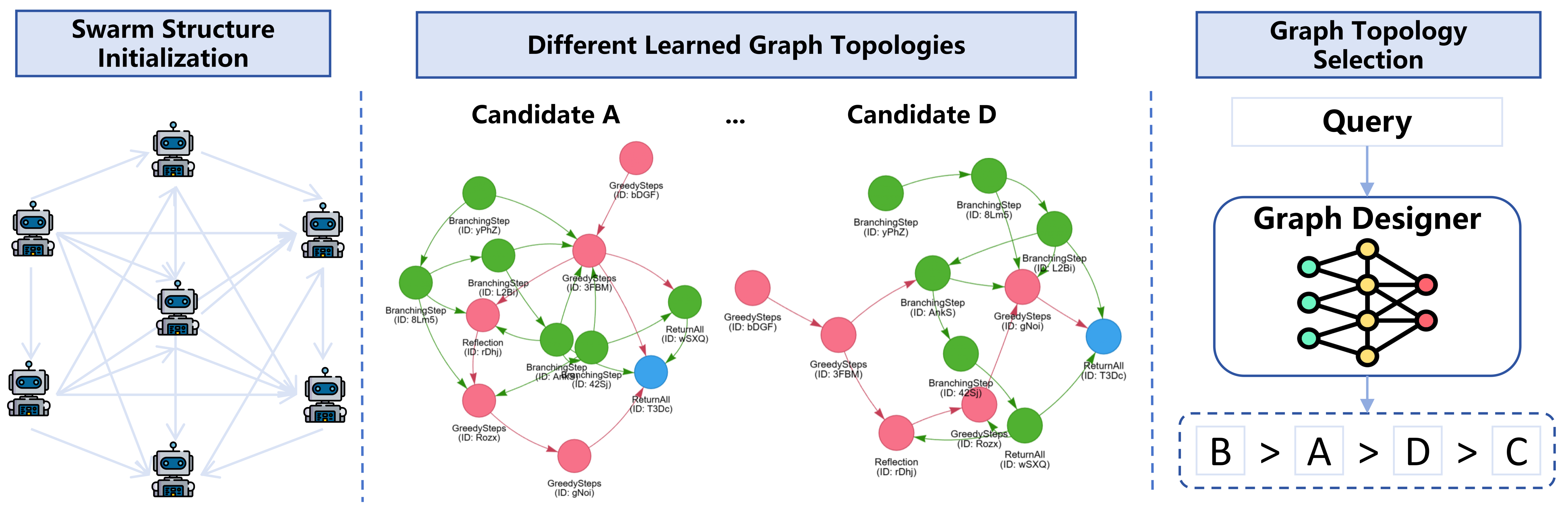}
\caption{Schematic illustration of our AMAS framework. }
\label{fig:architecture}
\end{figure*}

\section{Related works}

\subsection{LLM-based agents}

Large language models have undergone unprecedented advancements, exhibiting exceptional versatility across multifaceted application domains. Consequently, scholarly and industrial communities have intensified focus on transforming these models into autonomous cognitive agents. While LLM-driven single-agent systems demonstrate notable efficacy, the inherent advantages of collective intelligence remain irrefutable. Substantial research efforts have been directed toward LLM-powered multi-agent architectures. Drawing inspiration from the theoretical framework of collective cognition \cite{minsky1988society,zheng2024sca,zhang2024milora,zeng2025janusvlndecouplingsemanticsspatiality,zeng2025FSDrive,lu2022understanding,wang2025target,wang2024scantd,niu2025decoding,zhang2023moqagpt,wang2024coreinfer,wang2025anglesdontlieunlocking,liu2025qfft,liu2024rag,wang2025reasoningenhanceddomainadaptivepretrainingmultimodal}, NLSOMs \cite{zhuge2023mindstorms} deploy task-specialized social topologies within MAS implementations. The open-source ecosystem has witnessed proliferation of MAS development frameworks, including CAMEL \cite{li2023camel}, Agents \cite{zheng2024agentstudio}, ChatDev \cite{qian2307chatdev}, and AutoGen \cite{wu2023autogen}, which implement handcrafted role-assignment protocols for inter-agent coordination. MetaGPT \cite{hong2023metagpt} establishes structured operational frameworks to standardize role definitions and communication protocols, thereby enhancing collaborative efficiency. GPTSwarm \cite{zhuge2024gptswarm} conceptualizes MAS through composite topological architectures and proposes reinforcement learning-based parameterization for graph structure refinement. Despite these innovations, critical limitations persist: (a) automated topological optimization remains challenging due to reinforcement learning's inherent instability, and (b) all contemporary approaches enforce static graph configurations—whether manually engineered or RL-optimized—thereby neglecting sample-specific heterogeneity in task execution.

\subsection{Sample dependency in LLMs}

This research pioneers the integration of sample-specific heterogeneity into multi-agent system architecture, introducing a query-adaptive graph orchestrator that dynamically selects optimal topological configurations from candidate ensembles based on input characteristics. The conceptual foundation draws inspiration from parallel advancements in large language model research: in-context learning methodologies \cite{rubin2021learning,Li2023UnifiedDR} dynamically construct task-specific exemplars during inference to generate adaptive prompts, while input-dependent soft prompt tuning approaches \cite{zhu2024iapt,Liu2022LatePT} synthesize query-conditioned embedding vectors through parameter-efficient adaptation. AMAS extends this paradigm by transposing the input-aware design principle from prompt engineering to structural optimization, establishing a novel framework for context-sensitive multi-agent system architectures that fundamentally addresses sample-specific variation in task execution.

\section{AMAS}
\label{sec:method_AMAS}

\subsection{Preliminaries on graph optimization}

Drawing upon the theoretical foundation of collective cognition \citep{minsky1988society,zhuge2023mindstorms}, the GPTSwarm framework \cite{zhuge2024gptswarm} formalizes agent interconnectivity through a composite topological architecture $\mathcal{G} = (\mathcal{N}, \mathcal{E})$. To elevate multi-agent system efficacy, this approach further embeds topological attributes within differentiable parameters, employing policy gradient optimization via the REINFORCE algorithm to dynamically refine structural configurations for task-specific performance enhancement.

\subsection{A pilot experiments and motivations}

\begin{table*}
\centering
\resizebox{0.92\textwidth}{!}{
\begin{tabular}{c|ccccccccccccccc|c}
\hline

\multirow{2}*{Graph}&   \multicolumn{15}{c}{Sample index}    &     \\

   &    1 & 2 & 3 & 4 & 5 & 6 & 7 & 8 & 9 & 10 & 11 & 12 & 13 & 14 & 15  & Avg   \\ 

\hdashline

Graph A   &  0.2 &  0.2  &  
    0.1 &   
    0.2  &  
    0.3 &  
    0.4 &  
    0.1 &  
    0.1 &  
    0.2 &  
    0.3 &  
    0.2 &  
    0.3 &  
    0.1 &  
    0.1 &  
    0.3 &   0.208   \\
Graph B   &  0.1  &
    0.1  &
    0.2  &
    0.2  &
    0.2  &
    0.2  &
    0.3  &
    0.1  &
    0.1  &
    0.0  &
    0.2  &
    0.1  &
    0.2  &
    0.2  &
    0.2  &  0.199   \\
Graph C   &  0.1  &
    0.1  &
    0.1  &
    0.1  &
    0.1  &
    0.3  &
    0.0  &
    0.0  &
    0.2  &
    0.2  &
    0.3  &
    0.2  &
    0.5  &
    0.1  & 
    0.3  & 0.199   \\
Graph D   & 0.0  &
    0.1  &
    0.2  &
    0.0  &
    0.3  &
    0.1  &
    0.1  &
    0.3  &
    0.2  &
    0.2  &
    0.6  &
    0.2  &
    0.2  &
    0.0  &
    0.1  &   0.192   \\
\hline
\end{tabular}}

\caption{\label{tab:pilot_experiment_results} Pilot experiment's results on Crossword. This table presents the four different graphs' performances on 15 samples of the test set.} 
\vspace{-8pt}
\end{table*}

To establish the foundation for our AMAS framework, we initiate an exploratory investigation\footnote{The methodological approach aligns precisely with Section \ref{sec:experiments}, utilizing the Qwen2.5 3B language model as the core LLM component instead of the primary experimental configuration.} targeting the Crossword puzzle benchmark \cite{muralidharan2024deliberate}. Figure \ref{fig:architecture} displays the four most effective architectural designs—labeled Graph A through D—while Table \ref{tab:pilot_experiment_results} illustrates their evaluation outcomes across fifteen test instances. A complete tabular representation of these comparative results appears in Table \ref{tab:pilot_experiment_results}.

Analysis of the empirical findings uncovers two pivotal patterns: (i) While Graph A's architecture delivers the highest cumulative score in the Crossword evaluation, multiple alternative graph configurations demonstrate performance metrics that are statistically comparable to Graph A's outcomes. (ii) The assessment data reveals pronounced sample-specific performance variations, indicating that no single architectural design maintains consistent superiority across all test instances. Specifically, Graph A achieves the highest mean score yet fails to secure top position in every individual case. Conversely, Graph D registers the lowest average performance but never attains the lowest rank in any single evaluation. This fluctuation is exemplified by the contrasting performance hierarchies: the initial test instance ranks architectures as A > B = C > D, whereas the thirteenth sample exhibits a completely inverted ordering of C > B = D > A.

These empirical findings demonstrate that while reinforcement learning facilitates the refinement of agent architectures, a static graph configuration fails to secure consistent superiority across all task instances. Consequently, the integration of a dynamic graph selection mechanism—capable of autonomously evaluating and selecting the most appropriate architecture for each test sample based on predictive performance analytics—would yield a substantial performance enhancement for the agentic system.

\subsection{Construction of graph designer}

Our methodology for developing the graph designer within task $\mathcal{T}$ operates through a three-phase framework. Analogous to the reward modeling component in RLHF \cite{ouyang2022training}, this designer quantifies architectural efficacy by assigning a normalized performance expectation score within (0,1) for each candidate graph structure given the input context. Crucially, whereas RLHF reward models assess LLM output quality, our designer evaluates the intrinsic suitability of agent architecture configurations. The implementation pipeline proceeds as follows:

\noindent\textbf{Generation of Candidate Architectural Configurations} \quad The parameterized graph undergoes systematic refinement through optimization over task $\mathcal{T}$'s training corpus $\mathcal{D}_{train}$, executed according to the protocol specified in \cite{zhuge2024gptswarm}. This iterative process produces multiple parameter $\Theta$ checkpoint iterations, each yielding a distinct architectural configuration. Subsequently, we extract the top $K$ graph structures exhibiting maximal average performance metrics from these checkpoint-derived configurations.

\noindent \textbf{Formulating the Training Corpus for the Graph Selection Module} The graph designer's training dataset $\mathcal{D}_{gs, train}$ is systematically derived from $\mathcal{D}_{train}$ via a structured methodology. Each instance within $\mathcal{D}_{gs, train}$ operates on a dual-element architecture, featuring a query component $x$ and an outcome indicator $y$. The query $x$ follows a standardized template as follows:
\begin{lstlisting} 
Task Introduction:
(a) You are currently acting as the graph designer for the agent system that works on the [task_name] task. 
(b) The task [task_name]'s introduction is as follows: [task_intro].
(c) you will be given an input query, and a graph structure. Please evaluate the graph structure's quality in terms of how it will help solving the task in the input prompt. 

The input query is: 
[input_query].

The graph structure is:
[graph_structure]

\end{lstlisting}
In the above template, $\text{[task\_name]}$ denotes the the task name, $\text{[task\_intro]}$ denotes the introductory text contents for the task, $\text{[input\_query]}$ denotes the input query $q$ of the current sample, and $\text{[graph\_structure]}$ denotes the graph's structure $G$. And the label $y$ is the rank index for the $\text{[graph\_structure]}$. Correspondingly, $y$ encodes the sequential position of $\text{[graph\_structure]}$ within the structural hierarchy.

\noindent \textbf{Architecting the Graph Selection Mechanism} \quad The pre-existing LLM framework $\mathcal{M}$ serves as the foundational architecture, augmented via low-rank adaptation (LoRA) \cite{hu2021lora} to specialize in graph selection, driven by two critical advantages: (i) LoRA substantially reduces computational resource demands during training while mitigating reliance on extensive datasets; (ii) LoRA parameters integrate seamlessly with the existing LLM backbone, occupying merely $\sim$0.5\% of the backbone's GPU memory footprint. Let $\Omega$ denote the LoRA parameter set. The graph selection module integrates LoRA layers atop $\mathcal{M}$, supplemented by a pooling layer and a linear prediction head. Formally, with $\text{Pooler}(\cdot)$ representing the pooling operation and $\text{LP}(\cdot)$ denoting the prediction head, the designer's output is defined as:  
\begin{equation}
\hat{y} = \text{LP}\big(\text{Pooler}\big(\mathcal{M}(x \mid \Omega)\big)\big).
\end{equation}  
Here, $\mathcal{M}$ processes the input to generate hidden states $H_x \in \mathbf{R}^{l_x \times d_m}$. The $\text{Pooler}$ condenses these states into a contextual vector $h_x \in \mathbf{R}^{d_m}$, while $\text{LP}$ employs a linear transformation followed by a sigmoid activation to yield normalized scores in $[0,1]$.  

The training objective aligns the designer's predicted rankings with ground-truth performance metrics. For a test query $q$, $K$ candidate graphs $\{G_i\}_{i=1}^K$ correspond to distinct agent systems, each associated with performance score $s_i$. Their relative ordering is determined by:  
\begin{equation}
r_j = \text{Ranking}(s_j \mid \{s_j\}_{j=1}^K),
\end{equation}  
where $\text{Ranking}$ assigns positions $1$ (best) to $K$ (worst) in ascending score order. Ties are resolved by index priority (e.g., $i < j$ when $s_i = s_j$). \footnote{For identical scores $s_i = s_j$, the graph with smaller index $i$ receives higher rank.}  

To instill ranking semantics, we employ the loss:  
\begin{equation}
\mathcal{L}_{r} = \sum_{1 \leq i,j \leq K, i \neq j} m(i, j) \cdot g(i, j), \label{eq:loss_function}
\end{equation}  
with weights $m(i,j) = \max\big(0, |r_j - r_i|^{0.5}\big)$ \label{eq:weight_coefficient_in_loss} and scoring term $g(i,j) = \text{GeLU}\big((s_j - s_i) \cdot (\hat{y}_j - \hat{y}_i)\big)$.  

This formulation enforces critical constraints. When $s_j > s_i$ (implying $r_j > r_i$), $m(i,j) = 0$ discards the pair. When $s_j < s_i$ ($r_j < r_i$), $m(i,j) > 0$ and minimizing $\mathcal{L}_r$ maximizes $\hat{y}_j - \hat{y}_i$. Ties ($s_i = s_j$) yield $m(i,j) \cdot g(i,j) = 0$, rendering the pair inactive. The weight $m(i,j)$ dynamically scales loss contribution based on rank disparity. Adjacent ranks ($r_i=2$, $r_j=1$) yield $m(i,j) \approx 0.292$. Distant ranks ($r_i=4$, $r_j=1$) yield $m(i,j) = 0.5$, amplifying optimization pressure. Thus, $\mathcal{L}_r$ efficiently propagates list-wise ranking cues to the designer, enabling precise graph selection aligned with empirical performance.


\section{Experiments}
\label{sec:experiments}

\subsection{Datasets and evaluation metrics}

Our evaluation framework encompasses five rigorously designed assessment benchmarks: (i) \textit{Crossword} \cite{muralidharan2024deliberate}, requiring 5$\times$5 puzzle resolution; (ii) \textit{Game-of-24} \cite{muralidharan2024deliberate}, demanding arithmetic composition of four digits to reach 24; (iii) \textit{MMLU} \cite{hendrycks2020measuring}, a comprehensive multiple-choice reasoning benchmark; (iv) \textit{LLM-Eval-P} (internal benchmark), engineered to assess reasoning depth, factual knowledge, and task generalization across 47 domain-specific challenges spanning literature, healthcare, security, coding, and software engineering; (v) \textit{HumanEval} \cite{chen2021evaluating}, a code-generation evaluation suite. All datasets undergo standardized partitioning into 8:1:1 train/dev/test splits to support our AMAS pipeline. Graph designer fine-tuning data is exclusively derived from the training partitions, with comprehensive statistical profiles documented in Table \ref{tab:dataset_stats}.

\begin{table*}[tb!]
\centering
\resizebox{0.64\textwidth}{!}{
\begin{tabular}{cccccc}
\hline
Datasets  &  \#train    &  \#dev   &   \#test   &   Type   &  Metrics  \\ 
\hline

Game-of-24  &    1.0k   &  0.1k    &  0.1k    &      Math problem 
 &    acc     \\
Crossword   &    80   &  25  &  25    &      Text puzzles
 &    acc       \\

\hdashline

MMLU  &  11.2k  &  1.4k  &  1.4k   &    Question Answering    &    acc    \\

LLM-Eval-P  &  3.2k  &  0.4k  &  0.4k   &    Question Answering    &    acc    \\

\hdashline

HumanEval   &     120  &  24   &  20    &   Code generation   &  pass@10      \\

\hline
\end{tabular}}
\caption{\label{tab:dataset_stats} The statistics of the datasets evaluated in this work. }
\vspace{-8pt}
\end{table*}

Task-specific metrics are as follows: (i) \textit{Crossword} employs character-level precision, measuring the proportion of correctly resolved puzzle entries; (ii) \textit{Game-of-24} evaluates arithmetic composition success, quantifying the correctness of derived expressions from four numerical digits; (iii) \textit{MMLU} adopts multiple-choice reasoning accuracy, assessing selection correctness among candidate options; (iv) \textit{LLM-Eval-P} utilizes domain-diverse multiple-choice accuracy to gauge response correctness across 47 specialized challenge domains; (v) \textit{HumanEval} implements the standard \texttt{pass@10} metric, calculating the fraction of successful code executions across ten independent generation trials.

\subsection{Baselines}

We benchmark AMAS against state-of-the-art LLM architectures across diverse agent-centric inference paradigms. \textit{Monolithic agent approaches} encompass: (i) \textit{Input-Output} (IO), where the LLM directly synthesizes outputs from prompts; (ii) \textit{Chain-of-Thought} (COT) \cite{Wei2022ChainOT}, implementing stepwise reasoning prior to final response generation; (iii) \textit{Self-Consistency} \cite{wang2022self}; (iv) \textit{Tree-of-Thought} (TOT) \cite{muralidharan2024deliberate}; (v) \textit{Graph-of-Thought} (GOT) \cite{besta2024graph}. \textit{Collaborative agent ecosystems} include: (i) \textit{AutoGPT} \cite{yang2023auto}; (ii) \textit{AgentVerse} \cite{chen2023agentverse}; (iii) \textit{GPTSwarm} \cite{zhuge2024gptswarm}.

\subsection{Experiment Settings}
\label{subsec:experimental_settings}

\noindent\textbf{Computational infrastructure} \quad All experiments were conducted using either NVIDIA A40 GPUs (equipped with 48GB of memory) or NVIDIA A100 GPUs (featuring 80GB of memory).

\noindent\textbf{Foundation LLMs} \quad Each agent system in our study relies on a large language model (LLM) as its core backbone. Specifically, we employ the following models in our evaluations: (a) GPT-3.5-turbo\footnote{https://platform.openai.com/docs/models/gpt-3-5-turbo}; (b) the LLaMA-3 architecture \cite{dubey2024llama}, instantiated in both 8B and 70B parameter variants; and (c) distilled versions of the Deepseek R1 model \cite{guo2025deepseek}, each with 7B parameters.

\noindent\textbf{Graph optimization configuration} \quad Following the methodology of GPTSwarm, we structure our agentic system as a composite computational graph. This graph integrates three key components: a Tree-of-Thoughts (ToT) agent configured with a depth of 4 and a branching factor of 2; a Reflection agent \cite{shinn2024reflexion} that performs one reflection step over two iterative passes; and a dedicated output node. Altogether, this yields a graph comprising $n = 12$ nodes. The $d$ potential interconnections within the graph are governed by a learnable parameter vector $\Theta = [\theta_1, \theta_2, \dots, \theta_d]$. We utilize the REINFORCE algorithm \cite{williams1992simple}. On each optimization step, two graph structures are sampled via \cite{zhuge2024gptswarm}, and their will obtain rewards on the current batched samples. The parameters are optimized with the optimizer set to AdamW, the learning rate set to 1.0e-1, the training epoch set to 5, and the batch size set to 4. 

During optimization, we persist the graph parameters $\Theta$ at every tenth training step. Each saved checkpoint is then used to instantiate a concrete graph structure, which is subsequently evaluated on the development set. From these, we retain the top $K = 4$ highest-performing graphs as candidates for the downstream graph selection module.

\noindent\textbf{Graph designer hyperparameters} \quad Our implementation of the graph designer adopts the following settings: (a) the $\text{Pooler}$ utilizes last-token pooling—i.e., the representation of the final token in the input sequence serves as the aggregate embedding for the entire sequence; (b) a LoRA adapter with rank $r = 16$ is attached to every linear layer within the LLM backbone; and (c) for experiments involving proprietary LLMs, the designer is realized by fine-tuning the 7B distilled Deepseek model. In all other cases, the LoRA modules of the designer are fine-tuned directly on the same LLM backbone used by the agent. Consequently, when the backbone is the Deepseek 7B distilled model, the designer introduces an additional 40.5 million trainable parameters—equivalent to just 0.57\% of the total model size. At inference time, the designer assesses each candidate graph on a per-sample basis and selects the structure yielding the highest reward to construct the final agentic pipeline for prediction.

We use the HugginFace Transformers \cite{wolf-etal-2020-transformers} and PEFT \cite{peft} for implementing the training procedure of the graph designer. The batch size is set to ensure the optimization steps in one epoch is between 64 to 256, and the maximum training epoch is set to 10. We use AdamW as the optimizer with a linear learning rate decay schedule and 6\% of the training steps for warm-up. The learning rate is set to 1e-4. In every 50 steps, the model is evaluated on the dev set to calculate dev set perplexity. Patience is set to 10, that is, if the model does not achieve a lower dev set perplexity for 10 evaluation runs, the training stops early. The best checkpoint on the dev set is used to run predictions on the test set. 

\noindent\textbf{Reproducibility protocol} \quad To ensure robustness, every task is executed across five distinct random seeds, and we report the median performance across these runs.

\begin{table*}[tb!]
\centering
\resizebox{0.86\textwidth}{!}{
\renewcommand\arraystretch{1.1}
\begin{tabular}{c|ccccc|c}
\hline
\textbf{System}     &     \textbf{Crossword}    &     \textbf{Game-of-24}    &      \textbf{MMLU}      &     \textbf{LLM-Eval-P}      &   \textbf{HumanEval}    & \textbf{Latency}   \\ 
\hline

\multicolumn{7}{c}{\textbf{\emph{Results for LlaMA-3 8B}}}  \\
\hline

IO   &      0.165    &       0.132    &    0.536     &    0.365   &   0.659    &    1.13     \\
COT       &   0.184    &   0.217    &   0.594    &  0.432   &  0.701   &  2.25  \\
Self-consistency    &  0.205    &  0.206    &  0.604    &   0.445 
 &  0.707   &  2.67    \\
TOT     &   0.396      &  0.305     &   0.615   &    0.459    &   0.713   &  13.5  \\
GOT     &   0.406     &   0.289     &   0.618   &   0.464   &  0.708  & 14.6   \\

\hdashline

AutoGPT    &   0.418     &   0.309    &   0.621   &  0.457   &  0.698   &   32.4 \\
AgentVerse    &   0.452    &  0.326    &   0.632   &  0.458    &  
  0.715  &  35.2  \\
GPTSwarm    &    0.447      &  0.343    &   0.649    &   0.473  & 0.728    &   30.6 \\

\hdashline

AMAS (ours)   &    \textbf{0.485}    &    \textbf{0.377}     &  \textbf{0.663}    &   \textbf{0.481}     &  \textbf{0.748}   & 
 31.0 \\

\hline
\multicolumn{7}{c}{\textbf{\emph{Results for LlaMA-3 70B}}}  \\
\hline

TOT     &       0.647   &   0.521   &    0.829    &    0.564   &    0.785   &  145.6   \\
GPTSwarm    &  0.654   &    0.548   &   0.836   &   0.585   &  0.798   &   353.5 \\
\hdashline
AMAS (ours)   &    \textbf{0.671}   &   \textbf{0.563}   &    \textbf{0.847}   &   \textbf{0.597}  &   \textbf{0.812}       &  351.7 \\

\hline
\end{tabular}}

\caption{\label{tab:results_main_1} The Overall comparison of different agentic systems. The LLM backbone model is LlaMA-3 8B or 72B. We report the median accuracy over five random seeds. Bold indicate the best results.} 
\vspace{-2pt}
\end{table*}

\subsection{Main results}
\label{subsec:main_results}

We compare AMAS with baseline LLM agentic approaches, and the experimental results are presented in Table \ref{tab:results_main_1}. We present the average latency (in seconds) in the last column to examine the efficiency of each system. Table \ref{tab:results_main_1} reveals that: (a) our AMAS method outperforms the baseline methods across all seven tasks. In particular, AMAS outperforms the previous SOTA MAS baselines like AgentVerse and GPTSwarm. (b) Despite having an additional graph selection step, our AMAS's latency is comparable to that of GPTSwarm. The graph selection step requires only one forward pass on the LLM backbone, which will not significantly increase latency.

\subsection{Ablation studies and further analysis}
\label{subsec:ablation_studies}

\noindent\textbf{Results on more LLM backbones} \quad While our primary evaluation focuses on the open-source LLaMA-3 family, we further assess the generality of the AMAS framework by extending our experiments to a diverse set of language models: (a) GPT-3.5-turbo, (b) the distilled 7B variant of Deepseek R1, and (c) Qwen-3 models of both 8B and 30B scales \cite{yang2025qwen3}. Performance on the Crossword and Game-of-24 benchmarks is summarized in Table \ref{tab:results_different_backbones}. Due to practical limitations in integrating LoRA adapters with GPT-3.5-turbo, we instead train a graph designer by fine-tuning the LLaMA-3 8B model using LoRA. As shown in the table, our approach consistently surpasses conventional MAS baselines across these alternative backbones as well.

\begin{table}[tb!]
\centering
\resizebox{0.36\textwidth}{!}{
\begin{tabular}{c|cc}
\hline
\textbf{Method}   &     \textbf{Crossword}     &   \textbf{Game-of-24}      \\ 
\hline
\multicolumn{3}{c}{\textbf{\emph{Results for Deepseek R1 distilled 7B}}}  \\
\hline
TOT   &    0.457    &   0.368   \\
GPTSwarm       &   0.479   &   0.402    \\
\hdashline
AMAS    &    \textbf{0.516}     &   \textbf{0.438}   \\

\hline

\multicolumn{3}{c}{\textbf{\emph{Results for Qwen-3 8B}}}  \\
\hline
TOT   &    0.448    &   0.462   \\
GPTSwarm       &   0.464   &   0.471    \\
\hdashline
AMAS    &    \textbf{0.502}     &   \textbf{0.493}   \\

\hline
\multicolumn{3}{c}{\textbf{\emph{Results for Qwen-3 8B}}}  \\
\hline
TOT   &    0.592    &   583   \\
GPTSwarm       &   0.616   &   0.601    \\
\hdashline
AMAS    &    \textbf{0.642}     &   \textbf{0.615}   \\

\hline 
\multicolumn{3}{c}{\textbf{\emph{Results for GPT-3.5-turbo }}}  \\
\hline
TOT   &      0.673    &      0.646  \\
GPTSwarm     &  0.698     &   0.674  \\
\hdashline

AMAS    &    \textbf{0.717}     &   \textbf{0.692}   \\

\hline

\end{tabular}}
\caption{\label{tab:results_different_backbones} Experimental results for four different LLM backbones.}
\vspace{-8pt}
\end{table}

\noindent\textbf{Ablation analysis of the AMAS architecture} \quad To rigorously assess the architectural integrity of our AMAS framework, we systematically evaluate three distinct modifications: (a) AMAS-1, which employs $K = 8$ top-ranked graph candidates; (b) AMAS-2, which restricts candidate selection to $K = 2$ top graphs; (c) AMAS-3, which omits the weight coefficient $m(i, j)$ from the loss formulation (Equation \ref{eq:loss_function}). Comparative results across Crossword and Game-of-24 benchmarks are documented in Table \ref{tab:ablations}. Notably, the baseline AMAS configuration (mirroring Table \ref{tab:results_main_1}) achieves superior performance over all alternative implementations. Specifically: (a) AMAS-1 and AMAS-2 analyses confirm $K = 4$ as the optimal candidate threshold—reducing ($K=2$) or expanding ($K=8$) this parameter degrades graph designer efficacy. (b) The AMAS-3 comparison substantiates the loss objective's design (Equation \ref{eq:loss_function}), where $m(i, j)$ quantifies item-wise disparity ($i$ vs. $j$), thereby sharpening the model's sensitivity to ordinal relationships within the ranking structure.

\begin{table}[tb!]
\centering
\resizebox{0.34\textwidth}{!}{
\begin{tabular}{c|ccc}
\hline
\textbf{Method}   &     \textbf{Crossword}     &   \textbf{Game-of-24}      \\ 
\hline

AMAS   &     0.483    &     0.374     \\
\hdashline
AMAS-1   &   0.482    &     0.373    \\
AMAS-2   &   0.478    &     0.369      \\
AMAS-3   &    0.476   &   0.365     \\
\hline
\end{tabular}}

\caption{\label{tab:ablations} The comparison of AMAS's variants. } 
\vspace{-6pt}
\end{table}

\section{Conclusion}

This study introduces AMAS, a novel adaptive framework engineered to elevate LLM-driven multi-agent systems. We commence with an initial empirical investigation revealing task-specific sample sensitivity inherent in conventional MAS architectures. Subsequently, we engineer a graph designer mechanism that dynamically identifies optimal structural configurations for incoming queries. This designer is synthesized through parameter-efficient adaptation of the LLM backbone, leveraging our bespoke loss formulation. Comprehensive evaluations across question answering, mathematical reasoning, and code generation benchmarks affirm that AMAS delivers consistent superiority over leading single-agent and multi-agent baselines, across diverse LLM architectures. Crucially, AMAS achieves comparable computational efficiency to established approaches, establishing its viability for large-scale industrial deployment.

\section*{Limitations}

While our methodology demonstrates robust efficacy across diverse benchmarks and pretrained architectures, we recognize two key constraints: (a) computational constraints precluded evaluation on exceptionally large open-source LLMs, including LlaMA-3 450B and Deepseek R1. (b) The scope excludes more complex variants within mathematical reasoning, question answering, and information extraction domains. Notwithstanding these boundaries, the architectural adaptability of AMAS permits seamless integration with alternative backbone models and task paradigms. Future investigations will systematically examine the framework's performance across high-capacity model variants and challenging task landscapes, thereby validating its broader applicability beyond current experimental boundaries.

\section*{Ethics Statement}

This research establishes a paradigm for enhancing LLM-driven MAS architectures through optimized downstream performance. The experimental datasets represent established benchmarks in the literature, with comprehensive ethical clearance confirmed through peer-reviewed validation protocols. Our methodology was rigorously tested across LlaMA-3 variants, GPT-3.5-turbo, and Deepseek R1 distilled architectures. Notably, as with all generative language models, these systems exhibit inherent output unpredictability, occasionally generating erroneous or biased content. Crucially, this investigation centers on theoretical framework development for MAS methodologies, distinct from user-facing application deployment. Subsequent research will comprehensively evaluate the safety profile of AMAS within LLM operational ecosystems, prioritizing robustness against harmful outputs in future iterations.

\bibliography{custom}
\bibliographystyle{acl_natbib}

\end{CJK*}

\end{document}